\documentclass{article} 
\usepackage[final]{colm2026_conference}

\usepackage{microtype}
\usepackage{hyperref}
\usepackage{url}
\usepackage{booktabs}

\usepackage{wrapfig}

\usepackage{lineno}
\usepackage{graphicx}
\usepackage{amsmath}

\definecolor{darkblue}{rgb}{0, 0, 0.5}
\hypersetup{colorlinks=true, citecolor=darkblue, linkcolor=darkblue, urlcolor=darkblue}
\usepackage{booktabs} 
\usepackage{multirow} 
\usepackage[table,xcdraw]{xcolor}
\definecolor{gg}{HTML}{e2f0cb}
\definecolor{db}{RGB}{155, 189, 216}
\definecolor{think_blue}{RGB}{107, 146, 207}

\title{Can We Trust LLMs on Memristors? Diving into Reasoning Ability under Non-Ideality}

\author{Taiqiang~Wu\thanks{Equal contributions, $^{\dagger}$corresponding authors.}~, Yuxin~Cheng$^*$, Chenchen~Ding$^*$, \\ \textbf{Runming~Yang}, \textbf{Xincheng~Feng}, \textbf{Wenyong~Zhou}, \textbf{Zhengwu~Liu}$^{\dagger}$, \textbf{Ngai~Wong}$^{\dagger}$ \\
The University of Hong Kong, \texttt{takiwu@connect.hku.hk}
}

\begin{document}

\ifcolmsubmission
\linenumbers
\fi

\maketitle

\begin{abstract}
Memristor-based analog compute-in-memory (CIM) architectures provide a promising substrate for the efficient deployment of Large Language Models (LLMs), owing to superior energy efficiency and computational density.
However, these architectures suffer from precision issues caused by intrinsic non-idealities of memristors.
In this paper, we first conduct a comprehensive investigation into the impact of such typical non-idealities on LLM reasoning.
Empirical results indicate that reasoning capability decreases significantly but varies for distinct benchmarks.
Subsequently, we systematically appraise three training-free strategies, including thinking mode, in-context learning, and module redundancy.
We thus summarize valuable guidelines, i.e., shallow layer redundancy is particularly effective for improving robustness, thinking mode performs better under low noise levels but degrades at higher noise, and in-context learning reduces output length with a slight performance trade-off. 
Our findings offer new insights into LLM reasoning under non-ideality and practical strategies to improve robustness.
\end{abstract}

\section{Introduction}

Large language models (LLMs), such as GPT-4~\citep{openai_gpt4}, Llama~\citep{llama_report}, and Qwen~\citep{yang2025qwen3}, have demonstrated promising performance in Natural Language Processing (NLP) tasks, attributed to the superior reasoning capability~\citep{guo2025deepseek}.
However, this success, primarily driven by the massive model parameters and scaled chain-of-thought~(CoT) outputs, also presents challenges for practical deployments, such as high energy cost~\citep{alizadeh2024llm, wu2025halora}.


For efficient deployment of neural network, one promising approach is to leverage analog compute-in-memory (CIM) architectures with memristors such as resistive random-access memory (RRAM)~\citep{yao_nature}, which perform computations directly within memory arrays to achieve high energy efficiency and high computational density~\citep{li2024large}. 
By eliminating the energy-intensive data movement between memory and processing units, a key bottleneck in traditional computation system based on von Neumann architectures, CIM architectures are particularly well-suited for the matrix-vector multiplications that dominate LLM inference.

\begin{figure*}
    \centering
    \includegraphics[width=\linewidth]{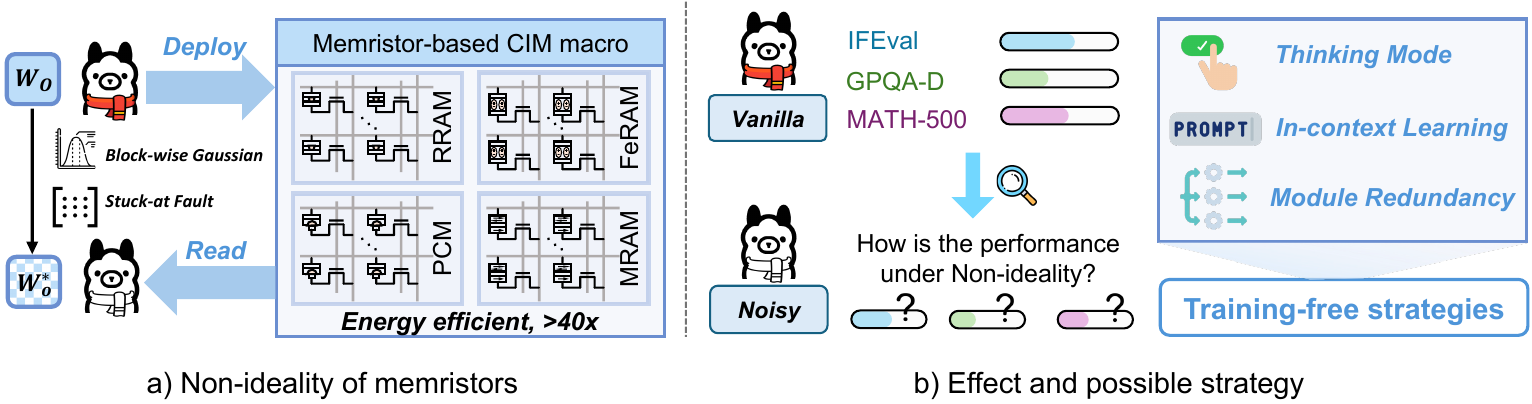}
    \caption{Overview of the non-ideality when deploying LLM on memristor-based analog CIM architectures.
    In this paper, we study the impact on LLM reasoning and discuss available training-free solutions.}
    \label{fig:overview}
\end{figure*}

This approach has also attracted increasing research attention for LLM deployment in recent years.
For instance, deploying Llama 3 1B and 3B on RRAM-based CIM saves more than 40$\times$ energy compared to GPU implementations~\citep{wu2025halora}.
However, despite its energy efficiency, memristor-based CIM suffers from inherent non-ideality issues, i.e., introducing noise during the calculation process.
As shown in Figure \ref{fig:overview}, this inherent non-ideality manifests as extra block-wise Gaussian noise and stuck-at faults in the weights.
Hence, this raises an intriguing question: \textit{How does such non-ideality impact the reasoning capability of LLMs? If so, can we eliminate or attenuate this side effect?}

In this paper, we conduct a comprehensive investigation into the impact of non-ideality in memristor-based analog CIM on LLM reasoning using RRAM-based CIM as an example.
Without loss of generality, we simulate multi-level non-ideality and then evaluate on three selected benchmarks: IFEval~\citep{zhou2023instruction}, GPQA-Diamond~\citep{rein2024gpqa}, and MATH-500~\citep{lightman2023let}.
Empirical results indicate that very slight noise~($\sigma=0.005$) can \textit{occasionally} lead to better results~\citep{wu2025halora}, while greater noise leads to severe performance degradation and greater variance.
In particular, the mathematical reasoning performance decreases rapidly, since they typically require multi-step reasoning and thus are more fragile to perturbations~\citep{wu2025revisiting}.
Meanwhile, the output tokens increase exponentially.
High noise would lead to excessively long outputs, nullifying the primary advantage of memristors.

To address this issue, re-training the LLMs is often unaffordable, as LLMs require huge amounts of data and thousands, or even millions, of GPU days~\citep{guo2025deepseek, yang2025qwen3}.
Hence, we systematically appraise three \textit{training-free} strategies to eliminate or attenuate the side effects of such non-ideality, including thinking mode, in-context learning~(ICL, \citep{dong2024survey}), and module redundancy~\citep{xia2017stuck}.
The thinking mode performs better under relatively low noise levels, but its efficacy diminishes at larger noise levels due to the collapse of the thinking mode.
ICL, by providing examples within the input, effectively reduces output length thanks to the provided extra solution patterns, but also suffers from a slight performance drop and increased input length.
For the module redundancy strategy, we find that greater redundancy generally improves robustness, and that shallow layers are particularly vital.
Based on these observations, we summarize practical guidelines including the behaviors and applicable scenarios.
In addition, we conduct experiments on more LLMs~(Qwen3 1.7B and Llama 3.2 1B) to demonstrate the effectiveness and robustness.

Our contributions can be summarized as follows:
\begin{itemize}
    \item We conduct a comprehensive investigation into the impact of memristor non-ideality on LLM reasoning, highlighting an important and non-negligible issue for practical deployment.
    \item We systematically appraise various training-free strategies to attenuate the impact of such non-ideality, and provide valuable insights into their mechanisms.
    \item We conduct extensive experiments on the Qwen3 and Llama series, and summarize practical guidelines for distinct scenarios.
\end{itemize}

\section{Related Work}
LLMs benefit from a consistently scaled reasoning trajectory~\citep{wei2022chain, guo2025deepseek}, driving rapid and significant advancements and enabling them to achieve state-of-the-art performance across various domains~\citep{ha2018world, ahn-etal-2024-large}.
With training for articulating long reasoning steps, the power of LLM is rapidly expanded to advanced mathematics~\citep{hendrycks2021measuring}, code generation~\citep{yu2018spider}, logical reasoning~\citep{ahn2022can, besta2024graph}, and autonomous planning applications~\citep{ferrag2025llm}. 
However, the substantial computational overhead associated with reasoning hinders the deployment and further application of advanced LLMs in edge computing~\citep{wang2025harnessing, feng2025efficient}. 

Consequently, research on integrating novel computing components, such as Memristors~\citep{wang2024enabling}, to optimize and adapt LLM has become increasingly important~\citep{vungarala2025limca, wu2025halora}.
Meanwhile, memristors achieve high energy efficiency but also suffer from non-ideality due to the stochastic nature of the switching process~\citep{joksas2020committee,chen2011variability}.
Previous works have studied the impact of non-ideality toward LLM finetuning~\citep{wu2025halora} and classification task via BERT~\citep{wang2024enabling, song2025hybrid}.

In this paper, we are the first to discuss the advanced LLM reasoning under non-ideality and discuss available training-free strategies.
\section{Reasoning Under Non-Ideality}

\subsection{Non-Ideality Simulation}

During inference, memristor non-ideality primarily manifests as random noise from device-to-device and cycle-to-cycle variability~\citep{neurosim_inf}.
Without loss of generality, we implement simulation for both \textit{block-wise} Gaussian noise and stuck-at faults.

\paragraph{Block-wise Gaussian Noise.}
This model introduces noise to the weights of individual memristor sub-blocks (tiles).
For a target memristor tile of shape $m\times n$, we partition the larger weight matrix $\mathbf{W}_0 \in \mathbb{R}^{d_1\times d_2}$ into smaller blocks.
The non-ideality simulation is then formulated as:
\begin{equation}
    \{\mathbf{W}_{0,[i,j]}\}_{i=1,j=1}^{k,t} = \operatorname{Split}(\mathbf{W}_0), \text{where} \ k=\lceil\frac{d_1}{m}\rceil, \ t=\lceil\frac{d_2}{n}\rceil,
\end{equation}
\begin{equation}
\begin{aligned}
\mathbf{W}_0^* = \operatorname{Cat}(\{\mathbf{W}_{0,[i,j]} + \mathbf{N}_{[i,j]} \cdot \max(|\mathbf{W}_{0,[i,j]}|)\}_{i=1,j=1}^{k,t})
\end{aligned}
\label{eq_rram_noise}
\end{equation}
\begin{equation}
\mathbf{N}_{[i,j]} \sim \mathcal{N}(\mathbf{0}, \sigma^2\mathbf{I})
\end{equation}
where $\operatorname{Split}(\cdot)$ and $\operatorname{Cat}(\cdot)$ denote the matrix splitting and concatenation operations. $\mathbf{W}_{0,[i,j]}$ is the partitioned weight block, and $\mathbf{W}_0^*$ is the resultant noise-injected weight matrix.
$\mathbf{N}_{[i,j]}$ is a noise matrix where each element is drawn from a Gaussian distribution with zero mean and variance $\sigma^2$.
The noise is scaled by the maximum absolute value of the block, and $\sigma$ represents the noise level, with a larger value denoting stronger noise.

\paragraph{Stuck-at Faults.}
Stuck-at Faults (SAFs) are physical defects where a memory cell becomes permanently stuck in either a high resistance state (HRS) or a low resistance state (LRS).
In this paper, we simulate the effect of SAFs by introducing random bit-flips to the mantissa of the bf16 weight values~\citep{chaudhuri2022functional}.
We model this as a probabilistic event: for each of the 7 bits in the mantissa, there is a probability $p$ that it will be flipped (0$\rightarrow$1 or 1$\rightarrow$0).
This parameter $p$ represents the SAF rate in the memristor array.
For example, consider the value 0.028:
\begin{equation}
0.028=0011110011100101
\end{equation}
If two bits happened to flip, then one possible outcome is:
\begin{equation}
0.029541015625=\underbrace{0}_{Sign}\underbrace{01111001}_{Exponent}\underbrace{11\textcolor{red}{1}0\textcolor{red}{0}01}_{Flipped \ Mantissa}.
\end{equation}
Then the read value is different from the input value.

\paragraph{Energy and Area Simulation.}
We assume that static vector-matrix multiplications are performed by memristor-based analog CIM (with RRAM used as the example).
Non-parametric operations like Softmax and activation operation are set to be computed in digital region. 
With this assumption, we developed a simulation system for energy and area estimation with RRAM-based analog CIM following the general practices applied in literature \citep{yao_nature,liu2025memristor}. 
The energy consumption and area of analog CIM macros is evaluated with XPESim \citep{xpesim}, and the digital region operations are evaluated with literature reported data \citep{tu_ISSCC}.

\subsection{Experimental Setup}
\label{sec:exp_setup}

Our primary experiments are conducted on Qwen3 0.6B~\citep{yang2025qwen3}.
Results on additional LLMs are presented in Section \ref{ana_more_llms}.
To simulate memristor non-ideality, we set the block size $m \times n$ to $64 \times 64$ and evaluate noise levels $\sigma \in \{0.005, 0.01, 0.015, 0.02\}$ following \citet{wu2025halora}.
For the Stuck-at Fault (SAF) simulation, the bit-flip probability $p$ is set to 0.01.
Under each noise configuration, we repeat 5 independent runs and report the mean score along with the standard deviation.

We evaluate performance on three prominent reasoning benchmarks: 1) IFEval~\citep{zhou2023instruction} for instruction-following ability;
2) GPQA-Diamond~(GPQA-D~\citep{rein2024gpqa}) for scientific reasoning ability using challenging multiple-choice questions from biology, physics, and chemistry;
and 3) MATH-500~\citep{lightman2023let} for mathematical reasoning using 500 uniformly selected test problems.
All benchmarks are evaluated in a zero-shot setting.
Besides the average accuracy~(Mean@8), we further report the Pass@8/Vote@8 and energy costs.
During generation, we follow the official guidelines for sampling parameters: temperature $t=0.7$, $\text{Top-p}=0.8$, and $\text{Top-k}=20$.
The maximum output length is set to 32k tokens, and we sample 8 responses for each prompt.
Evaluations are conducted on 8 NVIDIA H200 GPUs using the Opencompass framework~\citep{2023opencompass}.

\subsection{Main Results}

\begin{figure*}
    \centering
    \includegraphics[width=\linewidth]{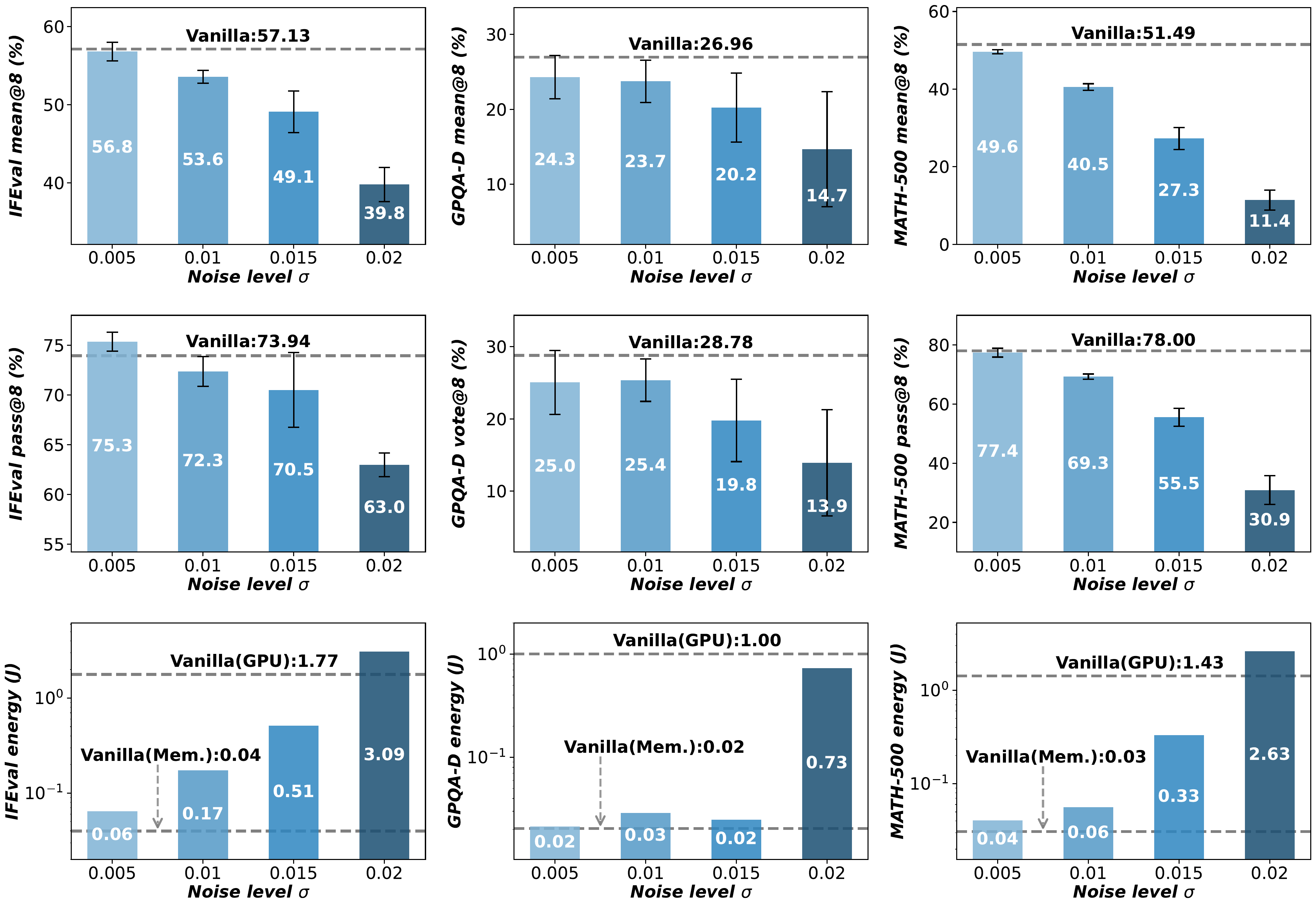}
    \caption{
    Performance of Qwen3 0.6B under Non-ideality at various noise levels $\sigma$.
    The SAF ratio $p$ is 0.01. 
    A darker color denotes larger noise.
    We report the results on IFEval, GPQA-D, and MATH-500.
    The vanilla performance is indicated by the dotted line.}
     \label{fig:exp_main}
\end{figure*}


Figure \ref{fig:exp_main} reports the main results on Qwen3 0.6B.
We can draw several key observations as follows.
\begin{itemize}

\item \textbf{Reasoning capability degrades under non-ideality, and instability increases with stronger noise.}
There is a strong negative correlation between the noise level $\sigma$ and reasoning performance.
As $\sigma$ increases from 0.005 to 0.02, performance on all six metrics degrades.
Furthermore, the standard deviation, visualized by the error bars, visibly increases with higher noise levels.

\item \textbf{Slight noise can sometimes lead to better performance.}
We observe a nuanced effect at the lowest noise level.
While most metrics are immediately degraded, the performance for IFEval Pass@8 at $\sigma=0.005$ (75.3\%) is slightly \textit{higher} than the ideal baseline (73.94\%).
This observation is consistent with the findings of \citet{wu2025halora}, suggesting that a minimal level of noise can sometimes act as a regularizer, potentially improving performance on specific tasks.
However, this minor benefit is selective and is quickly overwhelmed by the detrimental effects of noise, especially on MATH-500.

\item \textbf{Among these tasks, mathematical reasoning suffers most.}
Another key finding is that the impact of noise is highly task-dependent.
For mathematical reasoning (MATH-500), performance is extremely sensitive, with Pass@8 plummeting from 77.4\% down to 30.9\%.
This strongly suggests that tasks requiring high-precision, multi-step logical deduction are fundamentally more vulnerable to the weight perturbations caused by memristor non-ideality.

\item \textbf{The energy cost explodes under high noise, nullifying the primary advantage of memristor.}
The deployment on memristor is exceptionally energy-efficient, consuming only 0.03J on MATH-500 compared to the 1.43J of the GPU. 
However, this energy cost rises exponentially with the noise level, ballooning from 0.04J at $\sigma=0.005$ to a staggering 2.63J at $\sigma=0.02$. 
Critically, this 2.63J consumption at high noise is significantly worse than the 1.43J required by the standard GPU implementation, due to the excessively long and unstructured outputs.
Consequently, the intrinsic energy benefits of memristor are completely negated in high-noise scenarios.

\end{itemize}

\subsection{Error Analysis}

\begin{figure}[h]
    \centering   \includegraphics[width=0.85\linewidth]{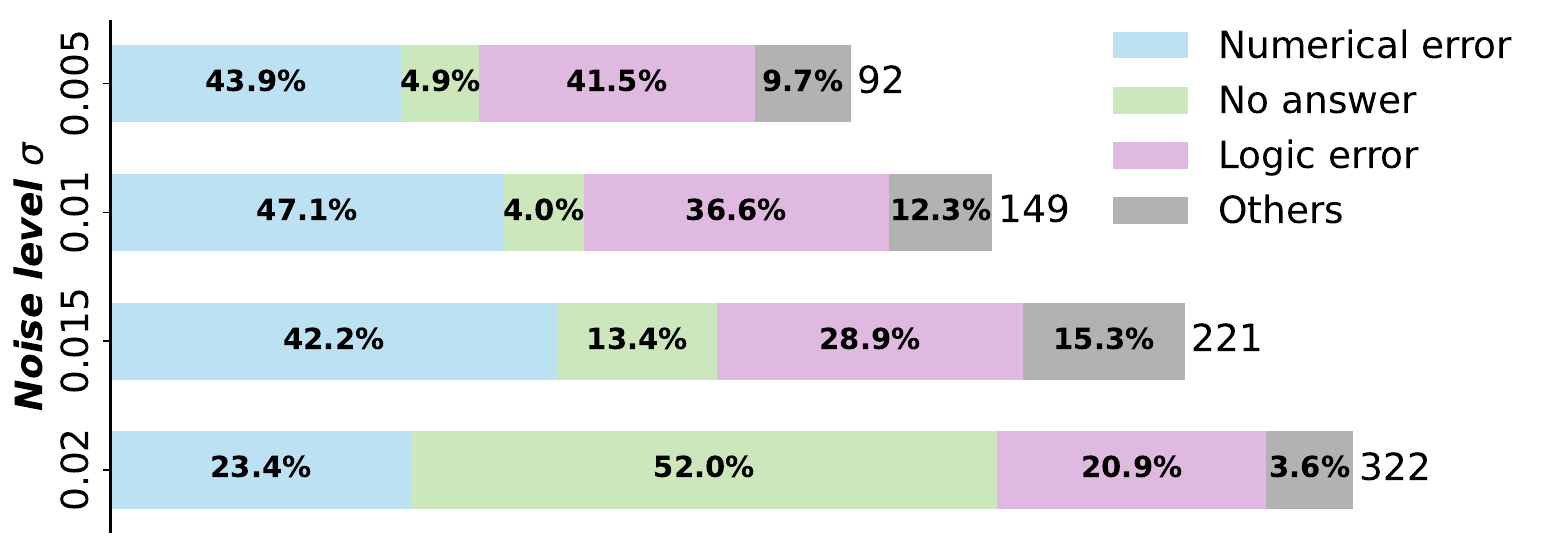}
    \caption{Error analysis of Qwen3 0.6B on Math-500.
    Larger noise leads to more formatting issues.}
    \label{fig:exp_error_analysis}
\end{figure}




We further analyze the answers from MATH-500 with $\sigma=0.02$ and employ Claude 4.5 to recognize the errors introduced by non-ideality.
The errors are categorized into four main types: 
1) \textit{Numerical Error} for calculating process error; 
2) \textit{No Answer} that fails to generate the final answer; 
3) \textit{Logic Error} where the reasoning process is coherent but flawed;
and 4) \textit{Others} for all other failure modes, such as formatting issues or incoherent text.
As shown in Figure \ref{fig:exp_error_analysis}, we observe that \textit{No Answer} increases as the noise becomes stronger.
The output reasoning trajectories mess up and fail to generate the final answer.
This suggests that as non-ideality worsens, the fundamental ability to adhere to the requested task structure collapses, leading to performance degeneration.

\subsection{Ablation on SAF ratio}

\begin{table}[h]
\centering
\begin{tabular}{lccc}
\toprule
 \multirow{2}{*}{\textbf{Noise}} & \textbf{IFEval} & \textbf{GPQA-D} & \textbf{MATH-500} \\
 & Mean@8 $\uparrow$ & Mean@8 $\uparrow$ & Mean@8 $\uparrow$\\
\midrule
Vanilla & 57.1 & 27.0 & 51.5 \\
\midrule
SAF~($p=0.0005$) & 57.0\scriptsize{$\pm$0.4} & 26.6\scriptsize{$\pm$1.1} & 51.5\scriptsize{$\pm$0.3} \\
SAF~($p=0.001$) & 57.0\scriptsize{$\pm$0.4} & 26.7\scriptsize{$\pm$1.2} & 51.6\scriptsize{$\pm$0.6} \\
SAF~($p=0.0025$) & 56.4\scriptsize{$\pm$0.9} & 26.7\scriptsize{$\pm$0.7} & 50.9\scriptsize{$\pm$1.0} \\
SAF~($p=0.005$) & 56.8\scriptsize{$\pm$0.4} & 25.4\scriptsize{$\pm$1.5} & 50.5\scriptsize{$\pm$0.9} \\
SAF~($p=0.01$) & 55.7\scriptsize{$\pm$0.7} & 26.1\scriptsize{$\pm$1.4} & 47.5\scriptsize{$\pm$1.5} \\
\bottomrule
\end{tabular}
\caption{Ablation of SAF ratio $p$.}
\label{tab:aba_p}
\end{table}

Moreover, we conduct an ablation study on the SAF ratio $p$.
As shown in Table \ref{tab:aba_p}, performance under the SAF only ($p=0.001$) setting (e.g., 57.0 on IFEval, 51.6 on MATH-500) remains close to the Vanilla baseline (57.1 on IFEval, 51.5 on MATH-500).
The larger the SAF ratio $p$, the worse the performance.
Typical value for the SAF ratio is about 0.005~\citep{guo2024hardware}.
Therefore, we fix $p$ to the highest typical value 0.01 in the following experiments.

\section{Extensive Analysis}

\subsection{Thinking Mode}

\begin{wrapfigure}{r}{0.65\textwidth} 
    \centering
    \vspace{-2em}
    \includegraphics[width=0.8\linewidth]{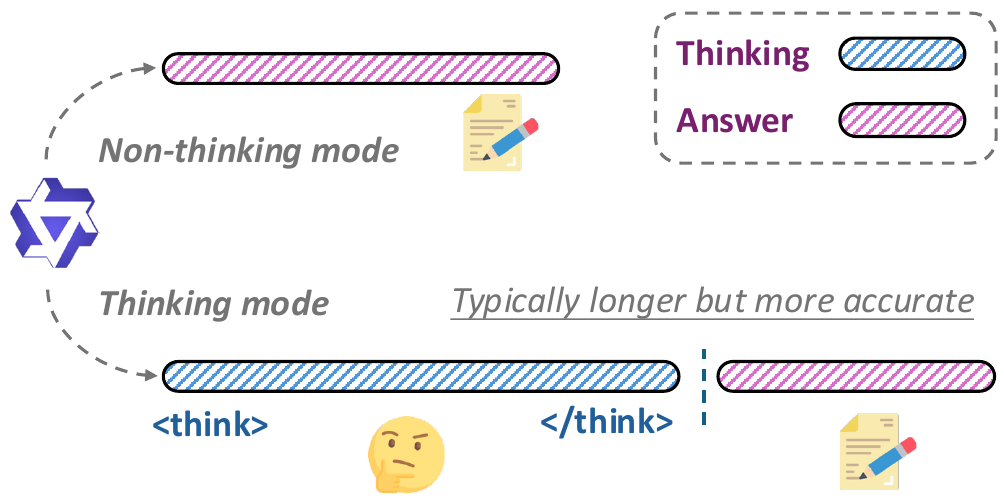} 
    \caption{Comparison between Non-thinking and Thinking mode that introduces an explicit thinking process.}
    \label{fig: thinking_toy}
\end{wrapfigure}

As shown in Figure \ref{fig: thinking_toy}, the thinking mode encourages the LLMs to explicitly output the thinking process before finalizing the answers.
Typically, the thinking mode outputs longer but more accurate responses.
Therefore, we turn on the thinking mode of Qwen3 models and set the hyperparameters accordingly: temperature $t=0.8$, $\text{Top-p}=0.95$.
Moreover, we report the token ratio of the Thinking parts among the outputs.

\begin{figure}[!h]
    \centering
    \includegraphics[width=\linewidth]{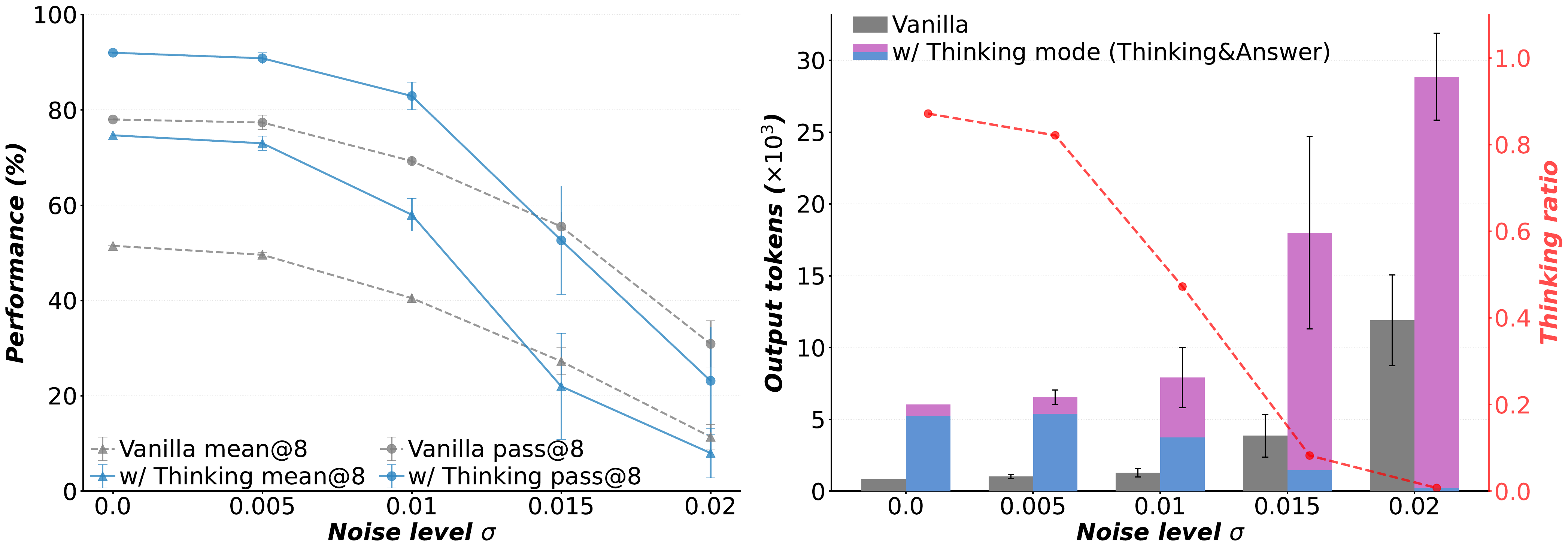}
    \caption{Comparison of Thinking and Non-thinking modes on MATH-500.
    The SAF ratio $p$ is 0.01.
    We also report the token ratio of \textcolor{think_blue}{Thinking} parts.}
    \label{fig:exp_cot}
\end{figure}

Figure \ref{fig:exp_cot} indicates the results of the MATH-500 benchmark on Qwen3 0.6B across the full range of noise levels. 
These results reveal that the effectiveness of the thinking mode is highly dependent on the noise level.
At relatively low noise levels ($\sigma \le 0.01$), the thinking mode provides a significant robustness advantage over the vanilla baseline.
However, for a larger noise level, the benefit of the thinking mode rapidly diminishes.
This failure is directly related to a mode collapse, as indicated by the thinking ratio plummeting to near-zero at $\sigma=0.02$.
Specifically, LLMs generate lots of \texttt{<think></think>} and repeat some phrases, such as \texttt{Therefore, the final answer is: xxx Therefore, the final answer is: xxx}.
It indicates that LLM fails to adhere to the explicit reasoning format and instead generates much longer but unstructured and repeated responses.

\textbf{Practical Guidelines.}
Based on the observations, we recommend the thinking mode as a viable training-free strategy only when the non-ideality is well-controlled~($\sigma \le 0.01$).
However, practitioners should be aware of two critical limitations.
First, the thinking mode incurs substantial computational overhead, generating 5-18$\times$ more output tokens than the vanilla model, leading to significantly higher inference latency and energy consumption on CIM architectures.
Second, the mode collapse phenomenon at high noise levels~($\sigma > 0.015$) renders this strategy not only ineffective but potentially harmful, as the model produces verbose yet unstructured outputs without meaningful reasoning.

\subsection{In-context Learning}


\begin{figure}[h]    
    \centering
    \includegraphics[width=\linewidth]{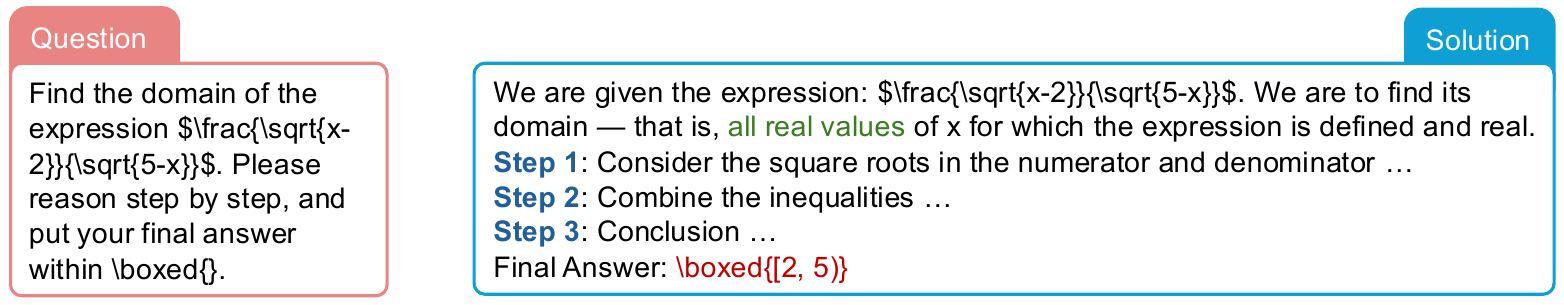}
    \caption{One example with a step-by-step solution under the in-context learning setting.
    We omit the details with ellipses.}
    \label{fig:icl_case}
\end{figure} 

Another strategy is in-context learning~(ICL), where LLM learns from a few examples in the context without updating the parameters~\citep{dong2024survey}.
To obtain high-quality examples, we adapt Qwen3-Max thinking to answer the given questions.
As shown in Figure \ref{fig:icl_case}, the response is well-organized and contains step-by-step solutions.
Since the answer is quite long, we employ the 2-shot setting to include two examples.

\begin{figure}[h]
    \centering
    \includegraphics[width=\linewidth]{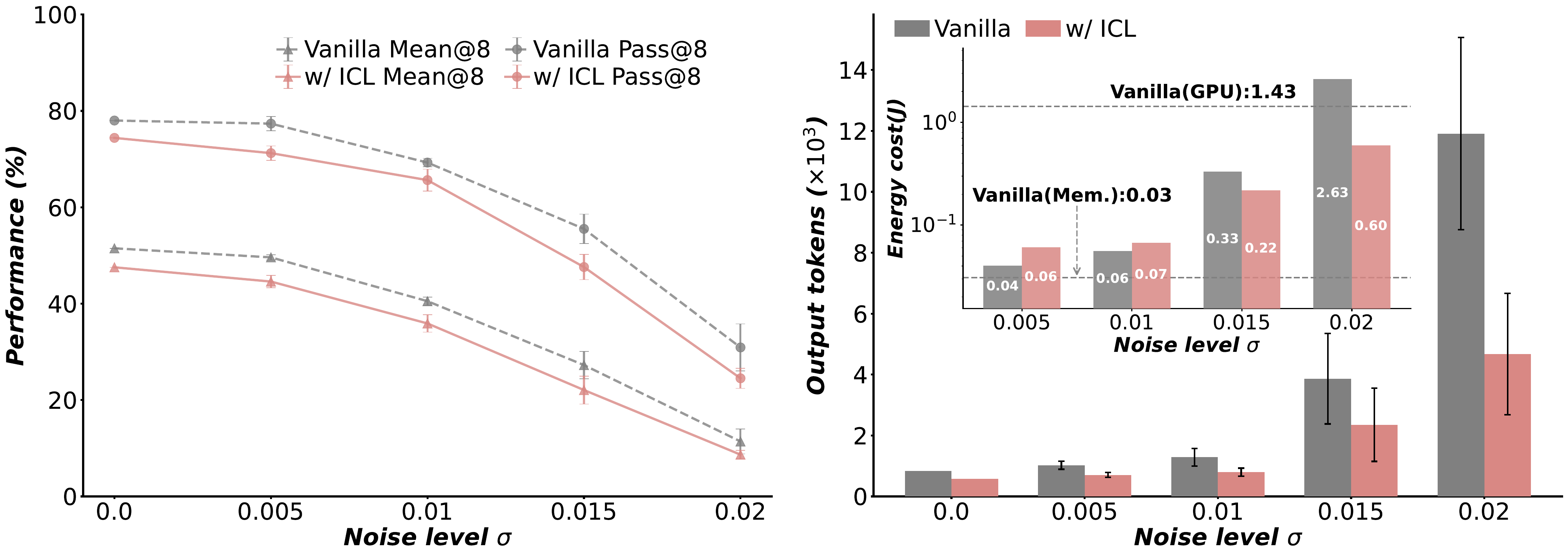}
    \vspace{-1em}
    \caption{Performance on MATH-500 with and without in-context learning (2-shots).
    The SAF ratio $p$ is 0.01.
    ICL effectively decreases the length of output tokens but also introduces longer input. 
    }
    \label{fig:exp_icl}
\end{figure}

Figure \ref{fig:exp_icl} indicates the detailed performance on MATH-500 for Qwen3 0.6B.
We observe that ICL consistently underperforms the vanilla model across all noise levels for both Mean@8 and Pass@8 metrics.
Meanwhile, ICL effectively decreases the length of output tokens compared to the exponential increase in the vanilla model.
However, ICL also requires longer inputs to provide the in-context examples, which introduces a critical trade-off. 
Therefore, the total energy consumption is not always lower. 
At low-to-moderate noise levels (e.g., $\sigma \le 0.015$), the energy cost of ICL is comparable to or even slightly higher than the vanilla RRAM baseline. 
The energy benefit of ICL only becomes apparent at very high noise levels ($\sigma = 0.02$), where it curtails the catastrophic token generation, resulting in significantly lower energy consumption (0.60J vs. 2.63J) at the cost of slightly lower accuracy.

\textbf{Practical Guidelines.}
ICL consistently leads to a degradation in reasoning accuracy (both Mean@8 and Pass@8) across all noise levels. 
At high noise levels, it acts as a safeguard against catastrophic energy consumption by enforcing a more structured and shorter output. 
However, in low-noise regimes, the longer input prompts required by ICL lead to slightly higher energy consumption while still delivering lower performance than the vanilla baseline. 
Future work could explore identifying more effective in-context examples to potentially improve robustness without this accuracy trade-off.

\subsection{Module Redundancy}

\begin{table*}[!t]
\centering
\resizebox{\textwidth}{!}
{
\begin{tabular}{lcccccccc}
\toprule
\multirow{2}{*}{\textbf{Method}} & \textbf{Area} & \textbf{Energy} & \multicolumn{2}{c}{\textbf{IFEval}} & \multicolumn{2}{c}{\textbf{GPQA-D}} &  \multicolumn{2}{c}{\textbf{MATH-500}} \\
\cmidrule(lr){4-5} \cmidrule(lr){6-7} \cmidrule(lr){8-9}
& (mm$^2$) $\downarrow$ & (J) $\downarrow$& Mean@8 $\uparrow$ & Pass@8 $\uparrow$ & Mean@8 $\uparrow$ & Vote@8 $\uparrow$ & Mean@8 $\uparrow$ & Pass@8 $\uparrow$ \\
\midrule
Vanilla (GPU)  & 806 & 1.43 & 57.1\scriptsize{$\pm$0.6} & 74.1\scriptsize{$\pm$0.3} & 27.0\scriptsize{$\pm$1.9} & 30.3\scriptsize{$\pm$2.1} & 51.5\scriptsize{$\pm$1.6} & 78.4\scriptsize{$\pm$0.6} \\
Vanilla (Memristor $\sigma$=0.02)  & 75 & 2.63 & 39.8\scriptsize{$\pm$2.2} & 63.0\scriptsize{$\pm$1.2} & 14.7\scriptsize{$\pm$7.7} & 13.9\scriptsize{$\pm$7.3} & 11.4\scriptsize{$\pm$2.6} & 30.9\scriptsize{$\pm$4.9} \\
\midrule
\multicolumn{9}{l}{\textit{Module Redundancy}} \\
\midrule
Attention ($\times$2) & 103 & 1.81 & 46.0\scriptsize{$\pm$3.8} & 70.5\scriptsize{$\pm$3.2} & 20.4\scriptsize{$\pm$5.4} & 19.8\scriptsize{$\pm$6.1} & 18.6\scriptsize{$\pm$2.9} & 41.3\scriptsize{$\pm$3.8} \\
Attention ($\times$4) & 160 & 0.45 & 48.3\scriptsize{$\pm$0.9} & 70.6\scriptsize{$\pm$3.2} & \cellcolor{gg}{20.7\scriptsize{$\pm$6.0}} & \cellcolor{gg}{21.4\scriptsize{$\pm$6.1}} & \cellcolor{db}{25.1\scriptsize{$\pm$2.8}} & \cellcolor{db}{51.9\scriptsize{$\pm$3.6}} \\
FFN ($\times$2) & 114 & 0.69 & \cellcolor{db}{49.6\scriptsize{$\pm$3.4}} & \cellcolor{db}{70.7\scriptsize{$\pm$3.2}} & 18.4\scriptsize{$\pm$9.0} & 18.3\scriptsize{$\pm$9.9} & 21.6\scriptsize{$\pm$2.5} & 47.6\scriptsize{$\pm$3.2} \\
FFN ($\times$4) & 193 & 0.22 & \cellcolor{gg}{52.2\scriptsize{$\pm$1.9}} & \cellcolor{gg}{73.2\scriptsize{$\pm$1.7}} & \cellcolor{db}{20.4\scriptsize{$\pm$3.5}} & \cellcolor{db}{19.2\scriptsize{$\pm$3.7}} & \cellcolor{gg}{30.1\scriptsize{$\pm$2.2}} & \cellcolor{gg}{58.7\scriptsize{$\pm$2.1}} \\
\midrule
\multicolumn{9}{l}{\textit{Layer Redundancy}} \\
\midrule
Layer 0-6 ($\times$2) & 91 & 1.94 & \cellcolor{gg}{45.1\scriptsize{$\pm$3.7}} & 67.4\scriptsize{$\pm$2.5} & 14.8\scriptsize{$\pm$3.6} & 13.8\scriptsize{$\pm$4.8} & \cellcolor{gg}{17.7\scriptsize{$\pm$0.9}} & \cellcolor{gg}{39.0\scriptsize{$\pm$0.6}} \\
Layer 7-13  ($\times$2) & 91 & 1.79 & \cellcolor{db}{44.7\scriptsize{$\pm$0.4}} & \cellcolor{gg}{69.7\scriptsize{$\pm$1.3}} & \cellcolor{db}{16.6\scriptsize{$\pm$7.9}} & \cellcolor{gg}{16.6\scriptsize{$\pm$9.0}} & 13.1\scriptsize{$\pm$4.1} & 34.0\scriptsize{$\pm$8.0} \\
Layer 14-20 ($\times$2) & 91 & 1.20 & 44.4\scriptsize{$\pm$3.3} & \cellcolor{db}{68.7\scriptsize{$\pm$2.2}} & \cellcolor{gg}{16.6\scriptsize{$\pm$5.0}} & \cellcolor{db}{16.2\scriptsize{$\pm$5.1}} & \cellcolor{db}{15.2\scriptsize{$\pm$1.3}} & \cellcolor{db}{36.7\scriptsize{$\pm$3.4}} \\
Layer 21-27 ($\times$2) & 91 & 4.58 & 42.2\scriptsize{$\pm$5.6} & 65.2\scriptsize{$\pm$5.9} & 10.0\scriptsize{$\pm$7.4} & 8.9\scriptsize{$\pm$7.5} & 10.9\scriptsize{$\pm$2.7} & 28.1\scriptsize{$\pm$4.7} \\
\bottomrule
\end{tabular}
}
\caption{Performance~(5 runs) of Qwen3 0.6B without and with module redundancy strategies. 
The Energy is calculated on the MATH-500 task.
For the noise simulation, $\sigma$ is 0.02 and $p$ is 0.01.
\colorbox{gg}{Green} denotes the best results, while \colorbox{db}{blue} for the second.}
\label{tab:redu_results}
\end{table*}

Another strategy is the module redundancy, aiming to reduce the noise level by repeating modules and calculating the average.
We investigate this by repeating the Attention and FFN modules, as well as entire blocks of layers, under a high noise condition ($\sigma=0.02$).

As shown in Table \ref{tab:redu_results}, repeating core components like FFN and Attention yields significant benefits. 
We observe that FFN redundancy is particularly effective. 
The FFN ($\times$4) strategy, which quadruples the FFN modules, achieves the best performance in this category, restoring the MATH-500 Mean@8 score from a baseline of 11.4 to 30.1 and Pass@8 from 30.9 to 58.7.
It also leads the IFEval recovery (52.2 for Mean@8). 
Attention ($\times$4) also provides a strong boost, securing the second-best results on MATH-500 (25.1 for Mean@8).
Crucially, this approach can also be exceptionally energy-efficient. 
While the noisy baseline consumes 2.63J, the FFN ($\times$4) strategy cuts this to just 0.22J, and Attention ($\times$4) to 0.45J. 
Both are far more efficient than the original GPU implementation (1.43 J), demonstrating that module redundancy can simultaneously recover reasoning performance and restore the energy advantage of memristor.

We also explored repeating entire layers. 
There are 28 layers in Qwen3 0.6B, and we repeat every 8 layers.
The results indicate that the early layers are most vital. Repeating the first 7 layers (Layer 0-6) provides a noticeable boost to MATH-500 performance (17.7 for Mean@8). 
However, repeating deeper layers is less effective or even detrimental; repeating the final layers (Layer 21-27) results in catastrophic energy consumption (4.58J) and poor performance. 
This suggests that robustness is highly dependent on the shallow or early layers of the model.

\textbf{Practical Guidelines.} 
Based on the results in Table \ref{tab:redu_results}, a clear trade-off emerges between performance, energy, and area. 
The observations suggest a targeted strategy applying heavy module redundancy only to the critical shallow layers.
This approach provides the most practical trade-off, capturing the significant performance and energy-efficiency gains observed in the 4$\times$ module tests while avoiding the prohibitive area cost of applying them to the entire model. 
This targeted method also avoids the severe performance and energy degradation seen when repeating deeper layers.

\subsection{Results on More LLMs}
\label{ana_more_llms}

We further conduct the experiments on Qwen3 1.7B and Llama 3.2 1B models.
Based on the guidelines, we try a simple strategy to repeat 4 times for the first 1/4 of the total layers, denoted as Shallow (4$\times$).
The decoding hyperparameters are set following official guidelines.
For the noise simulation, the $\sigma$ is 0.02.

\begin{table}[h]
\centering
\begin{tabular}{lccccc}
\toprule
\multirow{2}{*}{\textbf{Method}} & \textbf{IFEval} & \textbf{GPQA-D} & \textbf{MATH-500}  & \textbf{Energy} & \textbf{Area} \\
& Mean@8 $\uparrow$& Mean@8 $\uparrow$& Mean@8 $\uparrow$& (J) $\downarrow$& (mm$^2$) $\downarrow$\\
\midrule
\multicolumn{5}{l}{\textbf{Llama 3.2 1B}} \\
\midrule
Vanilla (GPU) & 50.0 & 23.2 & 15.8 & 12.0 & 806 \\
Vanilla ($\sigma=0.02$) & 39.9 & 17.4 & 7.6 & 1.5 & 121.3 \\
Shallow (4$\times$) & 47.4 & 18.7 & 8.5 & 0.7 & 207.1 \\
\midrule
\multicolumn{5}{l}{\textbf{Qwen3 1.7B}} \\
\midrule
Vanilla (GPU) & 67.1 & 29.9 & 74.0 & 4.6 & 806 \\
Vanilla ($\sigma=0.02$) & 40.9 & 10.5 & 19.0 & 7.2 & 174.3 \\
Shallow (4$\times$) & 59.9 & 26.7 & 53.8 & 0.3 & 299.9\\
\bottomrule
\end{tabular}
\caption{Performance on Llama 3.2 1B and Qwen3 1.7B.}
\label{tab:more_llms} 
\end{table}

Table \ref{tab:more_llms} reports the benchmark performance and energy on MATH-500.
The results clearly demonstrate that  Shallow (4$\times$) effectively achieves a good balance between performance and efficiency.
For Qwen3 1.7B, it recovers the MATH-500 score to 53.8 and IFEval to 59.9, closing a significant portion of the performance gap. 
Most importantly, it does so while being exceptionally energy-efficient, cutting the energy cost from 7.2J down to a mere 0.3J. 
The same trend holds for Llama 3.2 1B, where performance is recovered across all benchmarks (e.g., MATH-500 from 7.6 to 8.5) and energy is halved from 1.5J to 0.7J.
In summary, our guidelines are practical and applicable to other LLMs, demonstrating their effectiveness and robustness.

\section{Conclusion}

In this paper, we conducted a comprehensive investigation into the impact of non-ideality on LLM reasoning. 
Our findings reveal that reasoning capability, particularly for complex mathematical tasks, is highly vulnerable to hardware noise and degrades significantly as noise levels increase. 
Furthermore, this noise causes an exponential increase in output tokens, leading to severe energy consumption that can nullify the primary efficiency advantages of the memristor.
Therefore, we systematically appraise three training-free mitigation strategies and summarize corresponding guidelines.
Based on the observations, we further propose a practical strategy shallow~(4$\times$) that selectively repeats the first quarter of the model layers. 
Extensive experiments on more LLMs demonstrate that this approach achieves an excellent trade-off. 
Our findings offer valuable insights and practical guidelines for the robust deployment of LLMs on next-generation CIM hardware.

\section*{Ethics Statement}
We have adhered to the COLM Code of Ethics in this research. 
Our work is based entirely on publicly available models and benchmarks, and involves no human subjects.

\section*{Reproducibility Statement}

We are committed to the reproducibility of our work. 
Corresponding hyperparameters and detailed configuration for all experiments are documented in Section~\ref{sec:exp_setup}. 
All experiments were conducted on publicly available models and benchmarks.


\bibliographystyle{colm2026_conference}

\end{document}